\documentclass[12pt,a4paper]{article}
\usepackage[ansinew]{inputenc}
\usepackage[portuges,brazil,english]{babel}
\usepackage{model}
\usepackage{times}
\usepackage{epsfig}
\usepackage{graphicx}
\usepackage{amsmath}
\usepackage{amssymb}
\usepackage{color}
\usepackage{epstopdf}
\usepackage[pagebackref=true,breaklinks=true,letterpaper=true,colorlinks,bookmarks=false]{hyperref}
\usepackage{floatrow}
\DeclareFloatFont{footnotesize}{\footnotesize}
\cvprfinalcopy 

\ifcvprfinal\fi

\begin{document}

\title{Predict the model of a camera}
\author{Ciro J. Diaz Penedo}

\maketitle
\begin{abstract}
In this work we address the problem of predicting the model of a camera based on the content of their photographs. We use two set of features,  one set consist in properties extracted from a Discrete Wavelet Domain (DWD) obtained by applying a 4 level Fast Wavelet Decomposition  of the images as in \cite{Wang_2010}, and a second set are Local Binary Patterns (LBP) features from the after filter noise of images \cite{Huang_2011}. The algorithms used for classification were Logistic regression, K-NN and Artificial Neural Networks. 
\end{abstract}

\section{Introduction}
Digital cameras are today widely in use because of their good performance, convenient usability, and low costs.  The problem of how to recognize the
source camera of a given image has recently received a lot
of attention due to its appearance in many areas of work. \\
 
In our problem we have a  $2750$ photos set taken by $10$ different cameras ($275$ photos per each camera). There is an additional $2640$ photos set (unclassified) to  be classified by our model and results send to Kaggel to be evaluated. We will divide our data set in a Training Set $X_{tr}$ and a Cross validation Set $X_{cv}$. We extract a total $351$ features from wavelets coefficients in DWD \cite{Wang_2010} and with LBP we extract a total of $30$ features from the noise of our images. Principal Component Analysis (PCA) will be used to reduce dimensionality and  linear dependency. Data augmentation will be used to generate new samples to avoid over-fitting, bring more ``experience'' and better train our algorithms. \\  

  We propose the use of three different classification algorithms for source camera identification:   Logistic regression. K-NN and Neural networks, compare the results and come up with the best of them.

\section{Features extraction}

\subsection{Wavelet Features Extraction}

The two dimensional n-level discrete wavelet transform computes the coefficients of the expansion of an $L^2$ function (a matrix in the discrete case) in  orthonormal and compacted supported basis called wavelets \cite{DAUBECHIES_1992}. The coefficient can be separated in those corresponding to approximation at level $n$ and details (horizontal, vertical, and diagonal) at each level $\lbrace HV^{(k)}, HH^{(k)}, HD^{(k)}\rbrace_{k = n\cdots 1}$. \\

Following the steps in \cite{Wang_2010} we extract $3$ sets of $DB8$ wavelets  coefficients, $108$ features corresponding to statistics \textit{mean, variance, kurtosis and skewness} for both  details coefficients and linear predictor errors \cite{Lyu_2003} at each Level, color and band. In total we will have our first $216$ features.\\

So we also take the texture correlation existing in the wavelet coefficients into consideration. To do so we consider the co-occurrence matrix at each level, color  band and from it we extract the statistics Energy, Entropy, Contrast, Homogeneity and correlation as in \cite{HARALICK_1973} with let us another $135$ features. So our Wavelet Features Set (DWD features ) will have $351$ features. 

\subsection{LBP Features Extraction}
\label{FE}
Local binary patterns (LBP) is a type of visual descriptor used for classification in computer vision.  It has been found to be a powerful feature for texture classification.\\

We extract LBP features from the noise ($\mathcal{N}$) of grey images obtained by first filtering the original grey (Red, Green or Blue) image ($I$) to obtain ($I_F$) and then subtract it from the original image. The filtering of our images is done also via Wavelets, this time we use the type $bior3.5$ and put a threshold at some level of compression to keep different levels of noise (we only present results for threshold corresponding to the best classification).
\begin{equation}
\mathcal{N} = I - I_F
\end{equation}
LBP gives in general $59$ features where $49$ of them are associated to angle variations. In our case we are not interested in such modifications of images so we will keep only the remaining $10$ features for each color  and so our LBP Features set will have $30$ features.

\section{Features Selection}
\label{FS}
We are going to treat both sets of features by separate at each experiment. The idea is to keep a set of representative  features among the whole set of features. Some times features are highly correlated which means the influence of some of them in the model should be less important. For example, details coefficients in DWD are highly correlated \cite{Wang_2010}. This correlations may also cause an slow performance of the algorithms to minimize the cost function.\\

There are different methods to select features. One of them is PCA which  is a statistical procedure that uses an orthogonal transformation to convert a set of observations of possibly correlated variables into a set of values of linearly uncorrelated variables called principal components. We will use PCA to select new features from our DWD features  set. Keeping $87$ components will be enough to have a projection error $\mathcal{PE}$ 
of  $0.001$,
\begin{equation}
\mathcal{PE} = \Vert W - \mathcal{P}W \Vert_F = \sum_{k=71}^n \lambda_k
\end{equation}
where $\lambda_k$ are the eigenvalues of matrix $X_{tr} \cdot X_{tr}^T$.\\

Our LBP features set is already small so we wont need to select features from it, we will use it the way it is.

\section{Clasification}

To classify our data we will consider 3 methods Multinomial Logistic Regression (LR), K- NEarest NEighbour (K-NN),  Neural Network (NN) and compare their performances. We perform Feature extraction as in Section (\ref{FE}) to obtain sets DWD features  and LBP features. Then we do data augmentation and dimensional reduction for DWD features set as mentioned in Section (\ref{FS}) to apply K-NN and NN.  For LR we use the data set with no data augmentation.

\subsection{Multinomial Logistic Regression}
  This is a classification method that generalizes logistic regression to multi-class problems, i.e. with more than two possible discrete outcomes. The cost function to be minimized is.
  \begin{align*}
  \hspace{-1mm}
  J(\Theta) &= -\frac{1}{m}\left[ \sum_{i=1}^m\sum_{j=1}^k  \mathbf{1}\{y^{(i)}=j \} log \frac{exp( \Theta_j^Tx^{(i)})}{\sum_{1\leq l\leq k} exp( \Theta_l^Tx^{(i)})} \right] \\
  &+ \frac{\lambda}{2m} \sum_{j=1}^n \Theta_j^2 
  \end{align*}
where $k$ is the number of classes and $m$ is the number of samples and  $\Theta_j$ e $x^{(i)}$ are vector in $\mathbb{R}^n$ being $n$ the number of features. Unfortunately, there is no known closed-form way to estimate the parameters that minimize the cost function and thus we need to use an iterative algorithm such as gradient descent. The iterative algorithm requires us estimating the partial derivative of the cost function which is equal to
\begin{align*}
\frac{\partial J(\Theta)}{\partial \Theta_j} &= -\frac{1}{m}\sum_{i=1}^m\left[x^{(i)}\left( \mathbf{1}\{y^{(i)}=j \} - \frac{exp( \Theta_j^Tx^{(i)})}{\sum_{1\leq l\leq k} exp( \Theta_l^Tx^{(i)})}\right) \right] \\
&+ \frac{\lambda}{m} \sum_{j=1}^n \Theta_j 
\end{align*} 
We separate our set of $2750$ photos into a $80\%$ training set and $20\%$ Testing set. Applying this model to our data (Testing set of $55$ cameras in each class to be classified ) and selecting parameter $\lambda \in \lbrace 10^1, 10^0, 10^{-1},10^{-2}, 10^{-3}, 10^{-4}, 10^{-5}\rbrace $  we observe reduces the variance the most,  we obtain the confusion matrix for DWD features  set and LBP features  set shown in Tables (\ref{table1}) and (\ref{table2})
\begin{table}[h!]
\begin{tabular}{l |*{10}r |*1c}
            & C1 & C2 & C3 & C4 & C5 & C6 & C7 & C8 & C9 & C10 & \%\\
\hline
C1         &37 &    1 &    3 &   10 &   10 &    0 &    5 &    1  &   0 &  1 \\
C2         &2  &  48  &   4 &    1  &   0  &   1  &   0  &   1   &  1  &  0  \\
C3         &0  &   3  &  41  &   0  &   1  &   3  &   4  &   1   &  2  &  0 \\
C4         &9  &   0  &   0  &  36  &   0  &   1  &   4  &   0   &  0  &  0 \\
C5         &3  &   0  &   1  &   1  &  37  &   0  &   6  &   3   &  1  &  0 \\
C6         &1  &   1  &   2  &   1  &   2  &  49  &   0  &   0   &  0  &  0 \\
C7         &2  &   1  &   1  &   2  &   4  &   0  &  31  &   4   &  3  &  0 \\
C8         &0  &   0  &   1  &   2  &   1  &   0  &   3  &  42   &  6  &  0 \\
C9         &1  &   1  &   2  &   2  &   0  &   1  &   2  &   3   & 42  &  1 \\
C10        &0  &   0  &   0  &   0  &   0  &   0  &   0  &   0   &  0  & 53 \\
\hline
\%         & 58 & 89 & 63 & 75 & 69 & 75 & 49 & 64 & 76 & 96 & \textbf{73}$\%$\\
\end{tabular}
\caption{Prediction for Logistic regression using DWD features set. Mean percent of accuracy $73\%$}
\label{table1}
\end{table}
\begin{table}[h!]
\begin{tabular}{l |*{10}r |*1l}
            & C1 & C2 & C3 & C4 & C5 & C6 & C7 & C8 & C9 & C10 & \%\\
\hline
C1         &41 &    0  &   2 &    8 &    7 &    0  &   4&     0 &    0 &    0 \\
C2         &1  &  49   &  2  &   0  &   3  &   1   &  0 &    0  &   1  &   0  \\
C3         &1  &   2   & 45  &   2  &   1  &   0   &  1 &    0  &   0  &   0 \\
C4         &5  &   0   &  0  &  39  &   1  &   0   &  3 &    1  &   0  &   0 \\
C5         &3  &   0   &  2  &   1  &  36  &   0   &  7 &    2  &   1  &   0 \\
C6         &0  &   1   &  0  &   0  &   2  &  53   &  1 &    0  &   1  &   1 \\
C7         &0  &   0   &  1  &   4  &   4  &   0   & 33 &    2  &   2  &   0 \\
C8         &3  &   0   &  0  &   1  &   0  &   0   &  2 &   48  &   4  &   0 \\
C9         &1  &   3   &  3  &   0  &   1  &   1   &  4 &    2  &  46  &   0 \\
C10        &0  &   0   &  0  &   0  &   0  &   0   &  0 &    0  &   0  &  54 \\
\hline
\%         & 74 & 89 & 81 & 71 & 65 & 96 & 60 & 87 & 83 & 98 & \textbf{81}$\%$\\
\end{tabular}
\caption{Prediction for Logistic regression using LBP features set. Mean percent of accuracy $81\%$}
\label{table2}
\end{table}
  
  For Logistic regression the features LBP perform better than the $87$ principal components of DWD. Kaggle evaluation was $0.338124$ ($\textbf{33}\%$) for LBP and $0.330624$ ($\textbf{33}\%$) for DWD.
  
  \newpage
\subsection{K Nearest Neighbor (K-NN)}

 K-nearest neighbors algorithm (K-NN) is a non-supervised classification method. In k-NN classification an object is classified by a majority vote of its neighbours, with the object being assigned to the class most common among its k nearest neighbours.\\
 
 So let us usse here the data augmentation proposed in Subsection(\ref{NN}). Applying K-NN to our data (Testing set of $220$ cameras in each class to be classified) we obtain results shown in Tables (\ref{table3}) and (\ref{table4}) for LBP and DWD features
\begin{table}[h!]
\hspace*{-13mm}
\begin{tabular}{l |*{10}r |*1c}
            & C1 & C2 & C3 & C4 & C5 & C6 & C7 & C8 & C9 & C10 & \%\\
\hline
C1         &137 &    0 &   0 &    0 &    0 &    0 &    0 &    0   &   0 &  0 \\
C2         &0  &  180  &   0 &    0  &   0  &   0  &   0  &   0   &  0  &  0  \\
C3         &0  &   0  &   175 &   0  &   0  &   0  &   0  &   0   &  0  &  0 \\
C4         &0  &   0  &   0  &  178  &   0  &   0  &   0  &   0   &  0  &  0 \\
C5         &0  &   0  &   0  &   0  &  165  &   0  &   0  &   0   &  0  &  0 \\
C6         &0  &   0  &   0  &   0  &   0  &  193 &   0  &   0   &  0  &  0 \\
C7         &0  &   0  &   0  &   0  &   0  &   0  &  140 &   0   &  0  &  0 \\
C8         &0  &   0  &   0  &   0  &   0  &   0  &   0  &  197   &  0  &  0 \\
C9         &0  &   0  &   0  &   0  &   0  &   0  &   0  &   0   & 174  &  0 \\
C10        &0  &   0  &   0  &   0  &   0  &   0  &   0  &   0   &  0  & 216 \\
\hline
\%         & 62 & 81   & 79   & 80  & 75 & 87 & 63 & 89 & 79 & 98 & \textbf{73}$\%$\\
\end{tabular}
\hspace*{-13mm}
\caption{Prediction for K-NN using LBP features sets. Mean percent of accuracy $73\%$}
\label{table3}
\end{table}
\begin{table}[h!]
\hspace*{-30mm}
\begin{tabular}{l |*{10}r |*1c}
            & C1 & C2 & C3 & C4 & C5 & C6 & C7 & C8 & C9 & C10 & \%\\
\hline
C1         &113 &    0 &   0 &    0 &    0 &    0 &    0 &    0   &   0 &  0 \\
C2         &0   &  168  &   0 &    0  &   0  &   0  &   0  &   0   &  0  &  0  \\
C3         &0  &   0  &  185  &   0  &   0  &   0  &   0  &   0   &  0  &  0 \\
C4         &0  &   0  &   0  &  188  &   0  &   0  &   0  &   0   &  0  &  0 \\
C5         &0  &   0  &   0  &   0  &  180  &   0  &   0  &   0   &  0  &  0 \\
C6         &0  &   0  &   0  &   0  &   0  &  174  &   0  &   0   &  0  &  0 \\
C7         &0  &   0  &   0  &   0  &   0  &   0  &  125  &   0   &  0  &  0 \\
C8         &0  &   0  &   0  &   0  &   0  &   0  &   0  &  131   &  0  &  0 \\
C9         &0  &   0  &   0  &   0  &   0  &   0  &   0  &   0   & 147  &  0 \\
C10        &0  &   0  &   0  &   0  &   0  &   0  &   0  &   0   &  0  & 203 \\
\hline
\%         & 51 & 76 & 84 & 85 & 81 & 79 & 57 & 59 & 66 & 92 & \textbf{73}$\%$\\
\end{tabular}
\hspace*{-30mm}
\caption{Prediction for Logistic regression using DWD fetures sets. Mean percent of accuracy $81\%$}
\label{table4}
\end{table}
 
  In our case selecting $k = 8$ came up with the best classification with LBP features  and $k = 15$ with DWD features   the accuracy was $82\%$ for LBP features  $77\%$ for DWD features. Evaluation of our Test set classification in Kaggel gave $0.371249$ ($\textbf{37}\%$) for LBP and $0.296666$ ($\textbf{29}\%$) for DWD.

\subsection{Neural Networks}
\label{NN}
Artificial neural networks (ANN) is a supervised Machine Learning algorithm  inspired by the biological neural networks. Our  cost function in this case has the form
\begin{align*}
\hspace{-15mm}
J(\Theta) &= -\frac{1}{m}\left[ \sum_{i=1}^m \sum _{k=1}^K y_k^{(i)}log(h_\Theta(x^{(i)}))_k \right. \\
 &+ \left. (1-y^{(i)})log(1-h_\Theta(x^{(i)}))_k\right] \\
 &+ \frac{\lambda}{2m} \sum_{l=1}^{L-1} \sum_{i=1}^{s_l} \sum_{j=1}^{s_{l+1}} \left(\Theta_{ji}^{(l)}\right)^2
\end{align*}
where $K$ is the number of of unit (not counting bias unit) in the output layer, $h_\Theta(x) \in \mathbb{R}^K$ is our activation function and depends on all intermediate activations $\lbrace a^{(j)}_i \rbrace$ with $j = 2...L$ and $j = 1...s_i$, $L$ is the number of layers and $s_i$ the number of unit in the layer $i$. The computation of gradient of $J(\Theta)$ is done using the back propagation algorithm (BP). We do gradient check each some number of steps to verify that BP is doing Ok. Gradient descent algorithm is used to calculate zeros of $\triangledown J(\Theta)$.\\

The number of weights (parameters $\Theta^{(j)}_i$) can grow up fast as we increase the number of unit per layer and the number of layers, so It becomes necessary to do a data augmentation in order to train the Neural Network avoiding over-fitting. In our case we simply split each image into $5$ images, $4$ of them by \textit{spiting} each image from the center to each conner. Our set will have know $11000$ samples. We will keep $90\%$ for training, $10\%$ for validation.\\
  
  For LBP features set, we trained a NN with $1$ hidden layer and $60$ units, regularization parameter $\lambda$ was set to $7\cdot 10^{-5}$ where we saw less variance keeping more than $80\%$ of accuracy. It was not necessary to apply PCA to reduce features cause they were already a few ($30$).  For DWD features set, we trained a NN with $1$ hidden layer and $90$ units, regularization parameter $\lambda$ was set to $5\cdot 10^{-5}$ where we saw less variance while keeping more than $80\%$ of accuracy. We apply PCA to keep $87$ principal components from matrix $X_{tr}$ having in mind that was the number of features they used in (\cite{Wang_2010}), although they select features applying Sequential Forward Feature Selection (SFFS) but we expect that at least the not dependence will be preserved by using PCA (if not improved). \\
  
  The result of our classification experiments are presented via confusion matrix in  Tables (\ref{table5}) and (\ref{table6}). The classification accuracy was $77\%$ for DWD and $86\%$ for LBP and Kaggel  classification accuracy was $0.305416$ ($\textbf{31}\%$) for DWD and $0.412916$ ($\textbf{41}\%$) for LBP. 
\begin{table}[h!]
\hspace*{-13mm}
\begin{tabular}{l |*{10}r |*1c}
            & C1 & C2 & C3 & C4 & C5 & C6 & C7 & C8 & C9 & C10 & \%\\
\hline
C1         &59 &    0 &    4 &   3  &   5  &   0  &   8   &   9  &  4  &  1 \\
C2         &4  &  101 &   4  &   0  &   2  &   1  &   0   &   2  &  0  &  0  \\
C3         &7  &  5   &  71  &   2  &   1  &   0  &   5   &  2   &  7  &  0 \\
C4         &9  &   1  &   0  &  71  &   0  &   0  &   10  &  4   &  4  &  0   \\
C5         &6  &   5  &   3  &   3  & 103  &   0  &   3  &   4   &  5  &  0 \\
C6         &1  &   2  &   2  &   0  &   1  &  106 &   0  &   0   &  5  &  1 \\
C7         &5  &   1  &   2  &  10  &   2  &   0  &  78  &   5   &  4  &  0 \\
C8         &11 &   1  &  10  &   2  &   3  &   0  &   5  &  79   & 10  &  2 \\
C9         &8  &   1  &   0  &   2  &   1  &   2  &   2  &   6   & 75  &  2 \\
C10        &3  &   0  &   0  &   1  &   1  &   0  &   1  &   0   &  0  &109 \\
\hline
\%         & 52 & 86 & 74 & 75 & 86 & 97 & 69 & 71 & 65 & 94 & \textbf{77}$\%$\\
\end{tabular}
\hspace*{-13mm}
\caption{Prediction for NEural Network using $87$ principal components from DWD features set, $1$ Layer, $90$ units. Mean percent of accuracy $77\%$}
\label{table5}
\end{table}
\begin{table}[h!]
\hspace*{-13mm}
\begin{tabular}{l |*{10}r |*1c}
            & C1 & C2 & C3 & C4 & C5 & C6 & C7 & C8 & C9 & C10 & \%\\
\hline
C1         &87 &    1 &    1 &   3  &   5  &   0  &  10   &   0  &  0  &  1 \\
C2         &1  &  99  &   5 &   0  &   1  &   3  &   1    &   0  &  1  &  0  \\
C3         &2  &  2   &  99  &   2  &   3  &   1  &   1   &  1   &  0  &  0 \\
C4         &9  &   1  &   0  &  71  &   0  &   0  &   10  &  4   &  4  &  0   \\
C5         &6  &   1  &   3  &   7  & 108  &   1  &   6  &   0   &  0  &  0 \\
C6         &1  &   5  &   0  &   0  &   0  &   99 &   2  &   1   &  2  &  0 \\
C7         &7  &   2  &   0  &   1  &   6  &   1  &  75  &   2   &  1  &  0 \\
C8         &0 &   0  &    1  &   0  &   2  &   0  &   4  &  76   &  1  &  0 \\
C9         &1  &   0  &   0  &   1  &   1  &   2  &   5  &   4   & 110 &  1 \\
C10        &1  &   0  &   0  &   0  &   0  &   1  &   0  &   0   &  1  &109 \\
\hline
\%         & 80 & 89 & 87 & 87 & 82 & 91 & 68 & 84 & 95 & 98 & \textbf{86}$\%$\\
\end{tabular}
\hspace*{-13mm}
\caption{Prediction for NEural Network using $87$ principal components from LBP features set, $1$ Layer, $60$ units. Mean percent of accuracy $86\%$}
\label{table6}
\end{table}

\section{Conclusions}

  \textit{Samsung Galaxy Note 3} is the easy camera for model identification while the most difficult are \textit{Sony NEX-7} and \textit{HTC One M7}. The reason should be that \textit{Samsung Galaxy Note 3} takes photos with more resolution and definition and it Noise Pattern will be better defined as well, while the other two cameras are shipper with less definition and its Noise Patern will probably be affected more for external issues such light intensity. \\
  
  Our classification results in Kaggel where not that good. The best model result was Neural Network with LBP features set for which we got $41\%$ of accuracy. The reasons can be many. Maybe the features we extract are not sufficiently representative of the classes we want to classify. Maybe the models we are using to classify are not complex enough (Neural networks with more hidden layers, SVM and deep learning are other options can be explore). \\

One issue that without any doubt is affecting our result is the fact that we don't initially have images with the modifications of compression, resizing and gamma correction done to half of images in the Kaggel test set. Probably a data augmentation of our set with such modifications would improve any of the classification algorithms we use.\\

{\small
\bibliographystyle{unsrt}
\bibliography{Camera}
}

\end{document}